# Improved Parameter Identification Method Based on Moving Rate

Ho Chol Man
College of Computer Science ,
Kim Il Sung University,
Pyongyang, DPR of Korea
e-mail:
elib.rns@hotmail.com

Gwak Son Il
College of Computer Science ,
Kim Il Sung University,
Pyongyang, DPR of Korea

Pak Song Ho
College of Computer Science ,
Kim Il Sung University,
Pyongyang, DPR of Korea

Ha Jong Won
College of Computer Science ,
Kim Il Sung University,
Pyongyang, DPR of Korea
e-mail:
ryongnam2@yahoo.com

*Abstract*— **To improve the problem that the parameter identification for fuzzy neural network has many time complexities in calculating, an improved T-S fuzzy inference method and an parameter identification method for fuzzy neural network are proposed. It mainly includes three parts. First, improved fuzzy inference method based on production term for T-S Fuzzy model is explained. Then, compared with existing Sugeno fuzzy inference based on Compositional rules and type-distance fuzzy inference method, the proposed fuzzy inference algorithm has a less amount of complexity in calculating and the calculating process is simple. Next, a parameter identification method for FNN based on production inference is proposed. Finally, the proposed method is applied for the precipitation forecast and security situation prediction. Test results showed that the proposed method significantly improved the effectiveness of identification, reduced the learning order, time complexity and learning error.**

## I. Introduction

Fuzzy neural networks(FNNs) have become very popular over the past few years. An FNN is a hybrid method, which combines the semantic transparency of a rule-based fuzzy system with the learning capability of a neural network [1], [2], [3], [4]. The main advantage of FNN is that it resolves the black-box nature of a neural network. Fuzzy neural networks(FNNs) are an important paradigm in the development of hybrid intelligent systems for solving complex real-world problems such as pattern recognition, modeling, and forecasting. FNNs combine the learning, fault tolerant, and parallel processing abilities of neural networks, and the human inference and decision-making style of fuzzy inference systems.

Fuzzy neural networks are classified into two types. The first type is neural networks incorporating some fuzzy logic or fuzzy mathematic operation. The second type is neural networks based on fuzzy inference. The neural networks based on fuzzy inference are paid more attention. They include fuzzy neural networks based on Mamdani model and T–S (Takagi-Sugeno) model. For Mamdani model, the consequent parts of rules are a fuzzy set. For T-S model, the consequent parts of rules are functions of inputs. The main advantage of T–S model is to simulate the complex system with fewer rules [5].

For identification purposes, T-S models are also widely investigated, since they provide good interpolation and generalization characteristics. Identifying T-S fuzzy neural networks consists in two parts: structure modeling (determining the number of rules and the input variables involved), and parameters optimization or parameters identification (optimizing the consequent parameters) [6].

An improved fuzzy neural network based on Takagi-Sugeno(T–S) model is proposed in [5]. According to the properties of samples and the network structure, parameters are initialized to induce lower initial error. The results of simulations verify this method. In [7] presents a novel neural-fuzzy network architecture named evolving Mamdani - Takagi - Sugeno neural fuzzy inference system (eMTSFIS) that addresses two deficiencies faced by neural-fuzzy systems. The proposed eMTSFIS model combines Mamdani and T-S fuzzy modeling methods, coupled with a localized parameter learning method, to achieve both improved interpretability and accuracy. The paper [8] presents MTS-LiNFIS, a neural-fuzzy network that combines the Mamdani and T-S fuzzy modeling approaches to achieve improved interpretability-accuracy representation for linguistic fuzzy modeling.

In [9], a novel pattern classifying FNN based on the Yager inference rule with a novel DCTs was proposed. In addition, a modified Yager FNN was proposed, which improves the structure and learning algorithms of the POP-Yager FNN to give a better performance. A Tsukamoto-type Neural Fuzzy Inference Network (TNFIN) is proposed in [10]. This paper presented architecture of Tsukamoto-type neural fuzzy network and its associated hybrid learning algorithm.

The paper [11] proposes a new procedure for water level (or discharge) forecasting under uncertainty using artificial neural networks: uncertainty is expressed in the form of a fuzzy number. For this, the parameters of neural network, namely, the weights and biases are represented by fuzzy numbers rather than crisp numbers. Through the application of extension principle (Zadeh, 1965), the fuzzy number representative of the output variable is then calculated at each time step on the basis of a set of crisp inputs and fuzzy parameters of the neural network.

In most FNNs, only the identification phase of the parameters is determined using supervised or unsupervised - learning algorithms. The structure of these FNNs is fixed in

advance[12],[13],[14]; therefore, they are somewhat subjective—it is difficult for designers and technical domain experts to choose the appropriate number of rules to implement FNNs[15]. To solve this problem, self-organizing FNNs have been studied. Wu and Er proposed a hierarchical online self-organizing-learning algorithm for dynamic FNN (DFNN) based on work published in [16] and [17]. In [18], a new GP-FNN algorithm is introduced, which requires neither the number of neurons in the hidden layer nor the parameters to be predefined—the values are determined automatically during the learning process.

The identification method for pi-sigma fuzzy neural network based Sugeno method was proposed. But, its parameter learning method has many time complexities; because its learning algorithm uses the form of exponent function in fuzzy processing [19].

In this paper, a parameter identification method based on production fuzzy inference is proposed and its effectiveness is proved. The improved T-S fuzzy inference method is suggested for the case of which the consequent parts of rules are functions of inputs. The moving rate and production term is defined and the theoretical comparisons with Sugeno method and type-distance inference method are explained. Next, an improved identification algorithm (i.e. parameter learning) for pi-sigma fuzzy neural network based on production fuzzy inference method using efficient inference parameter is proposed. Using the production fuzzy inference method, the parameters learning of FNN for the experiment of forecasting of precipitation and security situation is studied on the PC and DE2 FPGA board, and by comparing with existing method, the efficiency of proposed method is proved.

## II. PRODUCTION INFERENCE METHOD USING EFFICIENT INFERENCE PARAMETER

In this section, the production inference method using efficient inference parameter is suggested and compared with existing inference method.

### A. Production inference method

For T-S Fuzzy model with n input and 1 output, the fuzzy system is presented as below.

$R_i$: if $x_1$ is $A_{i1}$ and $x_2$ is $A_{i2}$ and · · · and $x_j$ is $A_{ij}$
and · · · $x_n$ is $A_{in}$
then $y_i = c_{i0} + c_{i1}x_1 + \cdots + c_{ij}x_j + \cdots + c_{in}x_n$ (1)

$A_{ij}$ is $jth$ antecedent fuzzy set of $ith$ rules, $c_{i0}$, $c_{ij}$ are the coefficients of consequence straight line function and $i=1,\cdots,m$, $j=1,\cdots,n$. To explain production inference method using efficient inference parameter, the following concepts are defined as below for T-S fuzzy model, which is presented as formula (1) that has symmetry and asymmetry triangular membership function.

Firstly, the concept of moving rate is defined.

Fuzzy singleton for input (Fig. 1): if the center point of fuzzy set $A_{ij}$ is called $x_{cij}$, its right endpoint and left endpoint are called $x_{rij}$, $x_{lij}$, moving rate $d_{ij}$ for input information $x_{j0}$ can be got as followings:

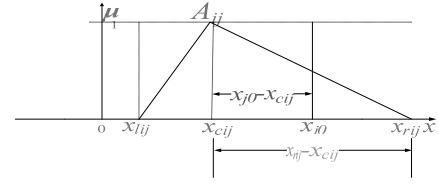

Figure 1. Calculation of moving rate with singleton input

$$d_{ij} = \begin{cases} \dfrac{x_{j0} - x_{cij}}{x_{rij} - x_{cij}}, & if \ x_{rij} > x_{j0} \geq x_{cij} \\ \dfrac{x_{cij} - x_{j0}}{x_{cij} - x_{lij}}, & if \ x_{lij} < x_{j0} \leq x_{cij} \\ 1, & if \ x_{j0} \leq x_{lij} \ or \ x_{rij} \leq x_{j0} \end{cases} \quad (2)$$

Fuzzy set for input（Fig. 2）: The moving rate is written as $d'_{ij}$ and got from expression (3). [Attention] The expressions (2) and (3) will be the same if the width of the fuzzy input information becomes 0. In other words, the expression (3) is generalized by (2).

$$d_{ij} = \begin{cases} \dfrac{x_{j0} - x_{cij}}{(x_{rij} - x_{cij}) + (x'_{rij} - x'_{cij})}, & if \ x_{rij} > x_{j0} \geq x_{cij} \\ \dfrac{x_{cij} - x_{j0}}{(x_{cij} - x_{lij}) + (x'_{cij} - x'_{lij})}, & if \ x_{lij} < x_{j0} \leq x_{cij} \\ 1, & if \ x_{j0} \leq x_{lij} \ or \ x_{rij} \leq x_{j0} \end{cases} \quad (3)$$

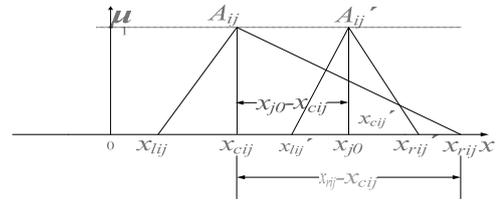

Figure 2. The calculation of moving rate with fuzzy input

Using the difference between input information $x_{j0}$ and central point $x_{cij}$ of antecedent fuzzy set $A_{ij}$, $d_i$ is called production term.

Using the moving rate $d_{ij}$ defined by expression (2), (3), production term $d_i$ is calculated as followings.

Fuzzy singleton for input: Production term $d_i$ is calculated as the expression (4).

$d_i = 1 - [d_{i1} \wedge d_{i2} \wedge \cdots \wedge d_{in}], \ i=1,\cdots,m, \ j=1,\cdots,n, \ if \ x_{j0} = \begin{cases} 1, & x = x_{j0} \\ 0, & x \neq x_{j0} \end{cases}$ (4)

Fuzzy set for input: Production term $d_i$ is calculated as the expression (5).

$d_i = 1 - [d_{i1} \wedge d_{i2} \wedge \cdots \wedge d_{in}], \ i=1,\cdots,m, \ j=1,\cdots,n, \ if \ x_{j0} = \mu_{Aij}(x_{j0})$ (5)

The production inference algorithm by using the above concepts is defined as follows.

*Step 1.* Calculate the moving rate $d_{ij}$ of center point of antecedent fuzzy set and input information with the expression (2), (3).

*Step 2.* Calculate the production term $d_i$ of $ith$ rule with the expression (4), (5).

*Step 3.* The number set $I_{act}$ of rules to attend the inference by efficient inference parameter $d_0$ is calculated as follows.

$$I_{act} = \{ i \mid d_i > d_0, 1 \leq i \leq m \} \qquad (6)$$

$$N = \mid I_{act} \mid , 1 \leq N \leq m \qquad (7)$$

$$d_o: \text{efficient parameter}, d_0 \in [0,1] \qquad (8)$$

*Step 4.* Finally, result $y_0$ is calculated by using the production term and the efficient rules, which are all got from above.

$$y_0 = \frac{\sum_{i=1}^{m} y_i \cdot d_i}{\sum_{i=1}^{m} d_i}, i = 1, \cdots, m \qquad (9)$$

For the fuzzy inference rules that is presented as expression (1), inference process using efficient inference parameter $d_0$ and production term that are defined by expression (4), (5) is called the production inference method or simple production method using efficient inference parameter.

*B. The comparison with the existing inference method*

*1) Basis for settings of moving rate, production term and efficient inference parameter:* First, the relationship between *Zadeh's* Compositional rules of inference and proposed production inference method is shown in Fig. 3.

In Fig. 3, we can know that the proposed moving rate $d_{ij}$ is equivalent to membership $\omega_{ij}$, production term $d_i$ to matching degree $u_i$ and efficient inference parameter $d_0$ to matching degree $\alpha_0$. And, it is possible to reason all whether the consequence of fuzzy set is fuzzy set or fuzzy singleton or straight line or state equation. For *sugeno* method, when the consequence is a constant or straight line [20], it is based on the compositional rules of inference.

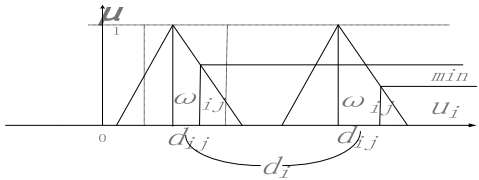

Figure 3. Relationship between production inference method and Compositional rules

Next, let's consider the basis of setting the moving rate. The proposed moving rate overcomes the disadvantage of existing move distance which considered only the difference between pure input information and fuzzy set. It can be said to be more reasonable because we talk about the difference considering the width of membership function.

Let us explain this with the relationship between memberships (10).

$$\text{Membership} \quad \omega_{ij} = \begin{cases} \dfrac{x_{rij} - x_{j0}}{x_{rij} - x_{cij}}, & \text{if } x_{rij} > x_{j0} \geq x_{cij} \\ \dfrac{x_{j0} - x_{lij}}{x_{cij} - x_{lij}}, & \text{if } x_{lij} < x_{j0} < x_{cij} \\ 0, & \text{if } x_{j0} \leq x_{lij} \text{ or } x_{rij} \leq x_{j0} \end{cases} \qquad (10)$$

From the expression (2) and (10), the more the matching degree of input information for fuzzy set is, the smaller the difference between input information and central point of fuzzy set, so the smaller moving rate.

In addition, the smaller the matching degree of input information for fuzzy set, the bigger the difference between input information and central point of fuzzy set, so the bigger the move rating. Matching degree plus the move rating will be one and it means that it has the basis of fuzzy set so as moving rate and matching degree. In the end, the moving rate is equivalent to matching degree and also has the other side. Matching degree shows the matching grade that input information belongs to fuzzy set by using membership function, but the moving rate shows the difference between input information and fuzzy set by using central point and width of fuzzy set. In other word, moving rate is not the concept to show matching degree of input information for fuzzy set, but is the concept to show the difference between input information and fuzzy set.

Next, production term is calculated using moving rate and shows matching degree between antecedent and input information. The production term is equivalent to the matching degree of compositional rule. The reason for setting the production term is to upgrade the usage of information so that antecedent fuzzy set can correctly apply for reason on incomplete input information. This is verified by comparison with existing method.

Next, on calculating, we can calculate by using addition and dividing. However, when it is studied using gradient method and addition is used, partial derivative function to other term except own is 0. Because division requires more instruction code than other does, it will cause CPU to peak word. So studying by production operation will be the effective way.

Efficient inference parameter is the concept that is equivalent to the threshold value of Compositional rule. If we look at setting basis of efficient inference parameter with the relationship of production term; term that is possible upgrading the usage of input information for the standpoint of rules is production term; when the effective rules among the rules that are possible upgrading the usage of input information for the standpoint of input information is selected, term that shows usage of rules is efficient inference parameter. In other word, when we deal with real problem, by using the efficient inference parameter that is equivalent to threshold value, rules that impact the results is participated to inference and it is removed rules that has less impact or is insufficient. As the value setting of efficient inference parameter will vary according to opponents by trial-and-error method.

*2) Comparison with sugeno method: Sugeno* method [20] is the fuzzy inference method that is applied to *T-S* fuzzy model based on Compositional rules of fuzzy inference. Here, we will calculate final output by calculating matching degree between antecedent membership and input information and weighted average of rules.

The common points of two methods are like this.

When antecedent is consisted of one input variable, production term $d_i$ is the same as matching degree $\omega_i$.

When $X_{li1} < X_{l0} \leq X_{ci1}$, $d_i = 1 - \dfrac{x_{ci1} - x_{l0}}{x_{ci1} - x_{li1}} = \dfrac{x_{l0} - x_{li1}}{x_{ci1} - x_{li1}} = \omega_i$.

When $X_{ci1} \leq X_{l0} < X_{ri1}$, it is elso calculated as same method.

When antecedent is consisted of many input variables, as production term is similar to matching degree, it will show uncertain matching between antecedent and input information. (Fig. 4)

The last output is calculated by weighted average. That is, the similar point of proposed method and *sugeno* method is that they calculate the matching degree and production term that show identical degree between input and antecedent fuzzy set, it is used by weight average in the calculation of final output.

The differences of these two methods are listed as below.

If input is not intersected with an antecedent fuzzy set of fuzzy rule $R_i$, matching degree is $\omega_i=0$ in *sugeno* method. Though input is not intersected with any antecedent fuzzy set in fuzzy inference method based on moving rate as expression (2), (3), it can take part in inference. For example, let's look at the fuzzy model that has 2 inputs and 1 output.

$$\text{If } x_1 = A_1, x_2 = A_2 \text{ then } y = c_0 \quad (11)$$

Consequence term will have constant term $c_0$.

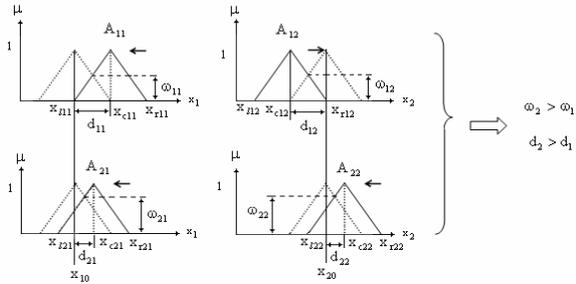

Figure 4.  Matching degree and production term between input and fuzzy set

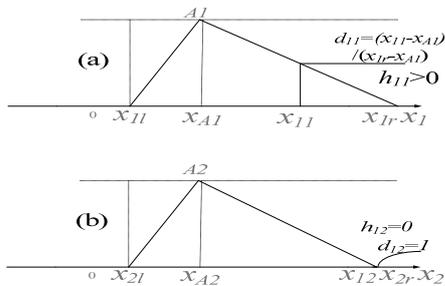

Figure 5.  Antecedent fuzzy set, input $x_{11}$ and $x_{12}$

In Fig. 5, the matching degree for $x_{11}$ is $h_{11} > 0$ (Fig. 5(a)) and the matching degree for $x_{12}$ is $h_{12} = 0$ (Fig. 5(b)). If entire matching degree is $\omega = h_{11} \wedge h_{12} = 0$, the consequence value of *sugeno* method is $y_0 = \omega \cdot c_0 /\omega = 0$. Therefore, the rule can not take part in fuzzy inference. But, if the production term is $d = 1-(x_{11} - x_{A1})/(x_{1r} - x_{A1}) > 0$, the result that is obtained by proposed method is bigger than 0. Therefore, it can take part in the fuzzy inference.

If any antecedent fuzzy set is not intersected with input information, proposed method obtains the production term. But, *sugeno*'s method can not be obtained final matching degree. That is, proposed method is similar for method that reasons with uncertain information; use possibly information than existing method. *Sugeno*'s method is inference method based on matching degree; proposed method is method based on moving rate and production term. That is, it is different for principle.

*3) Comparison with type-distance inference method:* The type-distance inference method [21], [22], [23] is derived the conclusion using distances.

For example, the fuzzy rule is presented as follows.

$R_i$: if $x_1$ is $A_{i1}$ and $x_2$ is $A_{i2}$ and ... and $x_j$ is $A_{ij}$ and ... $x_n$ is $A_{in}$

$$\text{then } y = f_i(x), i=1,...,m \quad (12)$$

Where, $x=(x_1,x_2,...,x_n)$ is antecedent variable vector, $A_{ij}$ is the antecedent fuzzy set, $f_i(x)$ is consequence function, $x_{jo}$ is input information, $x_0=(x_{10},x_{20},...,x_{n0})$ is input information vector and $y_0$ is inference result. In type-distance inference, distance function $d_i$ is calculated as follows (13).

$$d_i = \sum_{j=1}^{n} d(A_{ij}, A_{j0}) \quad (13)$$

When $x_{ij}$ is central coordinate of $A_{ij}$ and $x_{j0}$ is the central coordinate of $A_{j0}$, individual $n$ distance is calculated as $d(A_{ij}, A_{j0}) = |x_{ij} - x_{j0}|$. Inference result $y_0$ is calculated as expression (14). When $m=3, n=2$, expression (14) is like expression (15).

$$y_0 = \dfrac{\sum_{i=1}^{m}[f_i''(x_0) \cdot \prod_{j \neq 1} d_j]}{\sum_{i=1}^{m} \prod_{j \neq 1} d_j} \quad (14)$$

$$y_0 = \dfrac{f_1(x_0)d_2d_3 + f_2(x_0)d_1d_3 + f_3(x_0)d_1d_2}{d_2d_3 + d_1d_3 + d_1d_2} \quad (15)$$

The similarity between type-distance and proposed method is calculated the matching between antecedent fuzzy set and input information based on distance.

The differences between the two methods are as follows.

As we know in expression (15), even if input information is not all intersected with antecedent on type-distance inference, distances between rules and input or rules and rules are calculated. Therefore, it has irrelevant calculating for input. The type-distance has the problem that even if irrelevant input information with antecedent set is input, it is calculated in type-distance inference method. Such inference method is entirely contradiction to brain's thinking. If you receive any information, you do not decide any conclusion by using all brain's knowledge, not; you decide the conclusion by selecting proper knowledge. On the proposed method, if input information is not all intersected with antecedent fuzzy set, production term is $d_i=0$. Therefore, there is not partial inference result. In order to get result $y_0$ for type-distance inference, the expression (15) is used; $y_0$ for proposed method is calculated by expression (16).

$$y_0 = \frac{f_1(x_0)d_1 + f_2(x_0)d_2 + f_3(x_0)d_3}{d_1 + d_2 + d_3} \quad (16)$$

From expression (15) and (16), it can say that proposed method is logically more effective than existing method, has small complexity and is valid in practical application.

### III. THE IMPROVED IDENTIFICATION ALGORITHM FOR FUZZY NEURAL NETWORK

In this section, the identification algorithm (i.e. learning algorithm) for pi-sigma neural network using production inference method is described.

#### A. The pi-sigma neural network learning algorithm using sugeno method

If $ith$ rule is $R_i$, it is presented as follows.
$R_i$: if $x_1$ is $A_{i1}$ and $x_2$ is $A_{i2}$ and $\cdots$ and $x_j$ is $A_{ij}$
and $\cdots$ $x_n$ is $A_{in}$
then $y_i = c_{i0} + c_{i1}x_1 + \cdots + c_{ij}x_j + \cdots + c_{in}x_n$ (17)

$A_{ij}$ is a fuzzy set of $jth$ condition at $ith$ rule, $c_{i0}$, $c_{ij}$ is coefficient of consequence straight line function and $i=1,\cdots,m$, $j=1,\cdots,n$. The membership function of $A_{ij}$ is the Gaussian function $\mu_{Aij}(x_j)$.

The inference result $Y$ is calculated as follows.

$$y = \frac{\sum_{i=1}^{m} \mu_{A_{i1}}(x_1)\cdot\mu_{A_{i2}}(x_2)\cdots\mu_{A_{in}}(x_n))\cdot(c_{i0}+c_{i1}\cdot x_1+\cdots+c_{in}\cdot x_n)}{\sum_{i=1}^{m} \mu_{A_{i1}}(x_1)\cdot\mu_{A_{i2}}(x_2)\cdots\mu_{A_{in}}(x_n))} \quad (18)$$

$$= \frac{\sum_{i=1}^{m}(\omega_{i1}\cdot\omega_{i2}\cdots\omega_{in})\cdot(c_{i0}+c_{i1}\cdot x_1+\cdots+c_{in}\cdot x_n)}{\sum_{i=1}^{m}(\omega_{i1}\cdot\omega_{i2}\cdots\omega_{in})} = \frac{\sum_{i=1}^{m}\omega_i\cdot y_i}{\sum_{i=1}^{m}\omega_i}$$

The *pi-sigma* neural network [19] is a neural network that has addition neuron and production neuron as Fig. 6.

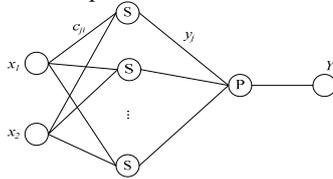

Figure 6.  pi-sigma neural network

Without loss of generality, it is supposed that neural network has *2* input neurons and *k* intermediate layer neurons. S is addition neuron and P is production neuron.

The output of neural network in Fig. 6 is

$$Y = \prod_{j=1}^{k} y_i = \prod_{j=1}^{k} \sum_{i=1}^{2} c_{ji} \cdot x_j \quad (19)$$

, $c_{ji}$ is the weight in neural network.

In practical problem, there are not only addition and multiplication operation but also fuzzy operation (max, min operation). Therefore, it is directly difficult to apply. So, as shown in Fig. 7, fuzzy neural network using *pi-sigma* neural network is discussed. In Fig. 6 and Fig. 7, S, P, · displays addition, multiplication, algebraic multiplication operation (or fuzzy operation min).

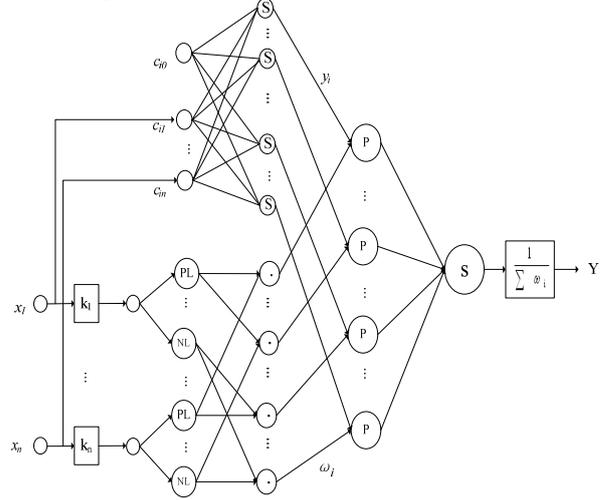

Figure 7.  Configuration of pi-sigma fuzzy neural network

As can be seen from the structure of neural network, network output is equal to expression (18). Therefore, for T-S model, inference method using this neural network is fully realized. This neural network is called *pi-sigma* fuzzy neural network. To discuss conveniently the learning of neural network, all membership functions of every fuzzy partial set are the Gaussian function. That is,

$$\mu_{A_{ij}}(x_j) = \exp\left[-(x_j - a_{ij})^2 / b_{ij}\right] \quad (20)$$

In fact, membership function to have more flexibility, any partial neural network can be replaced by Gaussian function. The goal output of neural network is $Y_d$, estimation function is defined as follows.

$$E = \frac{1}{2}(Y_d - Y)^2 \quad (21)$$

By gradient method, the learning is done as follows.

$$c_{ij}(t+1) = c_{ij}(t) - \eta\frac{\partial E}{\partial c_{ij}}\bigg|c_{ij} = c_{ij}(t) \quad (22)$$

$$a_{ij}(t+1) = a_{ij}(t) - \eta\frac{\partial E}{\partial a_{ij}}\bigg|a_{ij} = a_{ij}(t) \quad (23)$$

$$b_{ij}(t+1) = b_{ij}(t) - \eta\frac{\partial E}{\partial b_{ij}}\bigg|b_{ij} = b_{ij}(t) \quad (24)$$

#### B. Improved parameter learning algorithm based on production inference method

Using *pi-sigma* fuzzy neural network described above, for *T-S* fuzzy model, that antecedent is consisted of fuzzy set and consequent parts of rules are the functions of input variables, fuzzy neural network that realize the production inference method using effectiveness inference parameter is showed as Fig. 8. This fuzzy neural network fully realizes the production fuzzy inference and compared with *pi-sigma* fuzzy neural network, has the following differences.

First, algebraic production (·) neuron that calculates matching degree $u_i$ is to replace by $g$ neuron that calculates the production term $d_i$ reflected the difference between central point of fuzzy set and input information. Second, by adoption of the efficient inference parameter $d_0$, it is to display the excitement level of *P* neuron. It means that when fuzzy inference is done, the rules that are needed only participate to it.

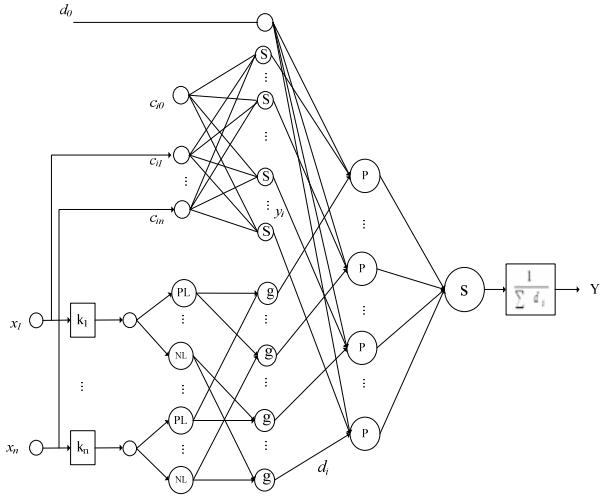

Figure 8. The configuration of pi-sigma fuzzy neural network using the efficient inference parameter

The learning of fuzzy neural network decides central point of antecedent membership function and the coefficient of consequence straight line function that error evaluation function expression (21) has the minimum value using gradient method as above. The expressions for error evaluation factions are as follows.

$$\frac{\partial E}{\partial x_{cij}} = \frac{\partial E}{\partial Y} \cdot \frac{\partial Y}{\partial d_i} \cdot \frac{\partial d_i}{\partial d_{ij}} \cdot \frac{\partial d_{ij}}{\partial x_{cij}} = \quad (25)$$

$$-(Y_d - Y) \cdot \frac{y_i \cdot \sum d_i - \sum y_i \cdot d_i}{(\sum d_i)^2} \cdot \prod_{k \neq j} d_{jk} \cdot \frac{(x_{sij} - x_j)}{(x_{sij} - x_{cij})^2}$$

$$\frac{\partial E}{\partial c_{i0}} = -(Y_d - Y)\frac{d_i}{\sum d_i} \quad (26)$$

$$\frac{\partial E}{\partial c_{ij}} = -(Y_d - Y)\frac{d_i}{\sum d_i}x_j \quad (27)$$

$$\Delta x_{c_{ij}}(t) = -\eta \frac{\partial E}{\partial x_{cij}}\Big|_{x_{cij}=x_{cij}(t)} \quad (28)$$

$$\Delta c_{i0}(t) = -\eta \frac{\partial E}{\partial c_{i0}}\Big|_{c_{i0}=c_{i0}(t)} \quad (29)$$

$$\Delta c_{ij}(t) = -\eta \frac{\partial E}{\partial c_{ij}}\Big|_{c_{ij}=c_{ij}(t)} \quad (30)$$

$$x_{ij}(t+1) = x_{ij}(t) + \Delta x_{ij}(t)$$
$$c_{i0}(t+1) = c_{i0}(t) + \Delta c_{i0}(t)$$
$$c_{ij}(t+1) = c_{ij}(t) + \Delta c_{ij}(t) \quad (31)$$

$x_{ij}(0)$ is the central point of fuzzy set $A_{ij}$, $c_{i0}(0)=c_{ij}(0)=0$ are coefficients of initialization consequence membership function , $t$ is a iteration test sequence and $t \in I_{act}$, $j=1,...,n$.

IV. COMPARISON WITH EXISTING METHODS BY LEARNING OF PRECIPITATION AND SECURITY SITUATION DATA

The proposed method is used in the test data learning of precipitation and security situation. And, compared with existing methods, its efficiency is proved.

The 36 fuzzy rules have been made out for the precipitation experiment by setting 6 fuzzy sets {*PL, PM, PS, NS, NM, NL*} (Fig. 9) to every input variable $x_j$ (*j*=1, 2) as (32).

R1: If x1=PL, x2=PL, x3=PL then c(0) + c(1)*x1 + c(2)*x2 + c(3)*x3
R2: If x1=PL, x2=PL, x3=PM then c(0) + c(1)*x1 + c(2)*x2 + c(3)*x3
...... (32)
R27: If x1=PS, x2=PS, x3=PS then c(0) + c(1)*x1 + c(2)*x2 + c(3)*x3

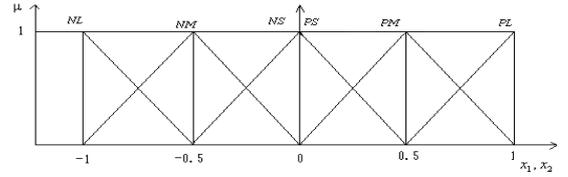

Figure 9. Membership function of $x_1$, $x_2$

The measured data from 1952 to 1977 by China Tianjin city Weather Service is shown in Table 1[24].

In the precipitation data, the inference results of *Sugeno* and the proposed method are as Fig. 10. Fig. 11 shows the comparison of learning time by order in two methods (t=32760). Table 2 shows the comparison results of average learning time and accuracy.

TABLE I. MEASURED DATA FOR PRECIPITATION FORECAST MODEL

| year | factor 1 | factor 2 | precipitation (mm) | year | factor 1 | factor 2 | precipitation (mm) |
|---|---|---|---|---|---|---|---|
| 1952 | 0.73 | -5.28 | 283 | 1965 | 0.46 | -14.68 | 348 |
| 1953 | -2.08 | 5.18 | 647 | 1966 | -2.31 | -1.36 | 644 |
| 1954 | -3.53 | 10.23 | 731 | 1967 | 0.2 | -5.43 | 431 |
| 1955 | -3.31 | 4.21 | 561 | 1968 | 3.46 | -19.85 | 179 |
| 1956 | 0.53 | -2.46 | 467 | 1969 | 0.08 | 8.59 | 615 |
| 1957 | 2.33 | 7.32 | 399 | 1970 | 1.46 | 7.26 | 433 |
| 1958 | -0.32 | -10.81 | 315 | 1971 | 0.24 | -1.1 | 401 |
| 1959 | -2.35 | 3.85 | 521 | 1972 | 0.89 | -16.94 | 206 |
| 1960 | -0.95 | 2.74 | 472 | 1973 | -0.5 | 10.46 | 639 |
| 1961 | -0.64 | 6.0 | 536 | 1974 | 2.15 | -10.06 | 418 |
| 1962 | 0.92 | 0.65 | 385 | 1975 | -0.89 | 12.11 | 570 |
| 1963 | 2.98 | -11.83 | 259 | 1976 | 1.4 | -6.26 | 415 |
| 1964 | -0.85 | -2.3 | 657 | 1977 | -0.59 | 7.15 | 796 |

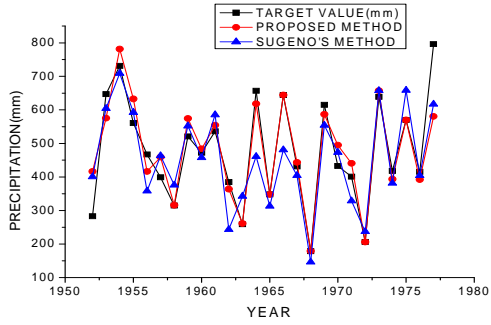

Figure 10. The experiment results of two method by year (t=32760)

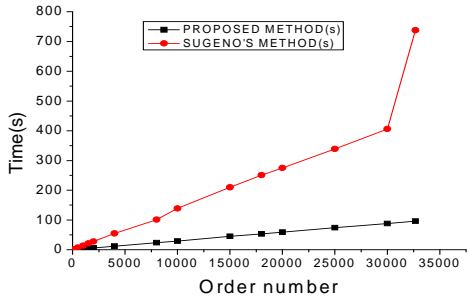

Figure 11. The comparison of learning time by order(s)

TABLE II. COMPARISON RESULT OF LEARNING TIME AND ACCURACY FOR TWO METHODS (T=32760)

| No | Learning method | time (s) | accuracy (%) |
|---|---|---|---|
| 1 | Proposed method | 96 | 92.09 |
| 2 | Sugeno's method | 738 | 84.74 |
| 3 | update | 7.68 times | 7.35 |

The prediction of network security situation evaluates the security situation by learning data for the past time and predicts the situation in the future. This method predicts and studies using fuzzy neural networks [25].

The security situation data is shown in Table 3. For a simple comparison experiment, *PL, PM* and *PS* are chosen and $x_1, x_2$ and $x_3$ are used as the input variable. $c_{i0}, c_{i1}, c_{i2}, c_{i3}$ are the coefficients of consequence function.

The fuzzy rules are as follows.

R1: If x1=PL, x2=PL, x3=PL then c(0) + c(1)*x1 + c(2)*x2 + c(3)*x3
R2: If x1=PL, x2=PL, x3=PM then c(0) + c(1)*x1 + c(2)*x2 + c(3)*x3
.....                                                                    (33)
R27: If x1=PS, x2=PS, x3=PS then c(0) + c(1)*x1 + c(2)*x2 + c(3)*x3

TABLE III. MEASURED DATA FOR SECURITY SITUATION PREDICTION MODEL

| Factor 1 | Factor 2 | Factor 3 | Security situation value | Factor 1 | Factor 2 | Factor 3 | Security situation value |
|---|---|---|---|---|---|---|---|
| 1 | 2 | 7 | 13.792 | 2 | 1 | 8 | 16.354 |
| 1 | 6 | 9 | 14.783 | 9 | 4 | 2 | 94.707 |
| 5 | 1 | 9 | 37.333 | 8 | 4 | 8 | 77.354 |
| 5 | 2 | 9 | 37.748 | 6 | 2 | 3 | 48.992 |
| 7 | 6 | 8 | 62.803 | 8 | 3 | 7 | 77.11 |
| 4 | 2 | 1 | 29.414 | 6 | 4 | 5 | 49.447 |
| 8 | 6 | 6 | 77.858 | 9 | 4 | 6 | 94.408 |
| 6 | 9 | 3 | 50.577 | 4 | 1 | 6 | 28.408 |
| 4 | 2 | 6 | 28.822 | 4 | 8 | 4 | 30.328 |
| 5 | 2 | 1 | 38.414 | 2 | 6 | 7 | 17.827 |
| 1 | 1 | 4 | 13.5 | 2 | 4 | 9 | 17.333 |
| 3 | 4 | 7 | 22.378 | 1 | 3 | 7 | 14.11 |
| 1 | 5 | 2 | 14.943 | 5 | 6 | 7 | 38.827 |
| 1 | 8 | 9 | 15.162 | 2 | 7 | 8 | 17.999 |
| 9 | 1 | 1 | 94 | 7 | 8 | 6 | 63.237 |
| 9 | 3 | 7 | 94.11 | 4 | 9 | 6 | 30.408 |
| 3 | 9 | 9 | 23.333 | 1 | 4 | 6 | 14.408 |
| 4 | 7 | 5 | 30.093 | 3 | 9 | 1 | 24 |
| 3 | 4 | 9 | 22.333 | 3 | 7 | 6 | 23.054 |
| 5 | 7 | 1 | 39.646 | 1 | 7 | 3 | 15.223 |
| 4 | 7 | 7 | 30.024 | 1 | 1 | 1 | 14 |
| 7 | 8 | 9 | 63.162 | 5 | 7 | 7 | 39.024 |
| 2 | 4 | 4 | 17.5 | 8 | 1 | 7 | 76.378 |
| 1 | 8 | 4 | 15.328 | 4 | 6 | 8 | 29.803 |
| 6 | 7 | 1 | 50.646 | 2 | 4 | 7 | 17.378 |
| 4 | 1 | 1 | 29 | 7 | 4 | 8 | 62.354 |
| 9 | 4 | 1 | 95 | 1 | 4 | 1 | 15 |
| 6 | 2 | 1 | 49.414 | 6 | 6 | 5 | 49.897 |
| 3 | 1 | 8 | 21.354 | 4 | 1 | 9 | 28.333 |
| 5 | 4 | 5 | 38.447 | 3 | 2 | 4 | 21.914 |

In the security situation data, the inference results of *Sugeno* and the proposed method are as Fig. 12. Fig. 13 shows the comparison of learning time by order for two methods (t=32760). Table 4 shows the comparison results of average learning time and accuracy.

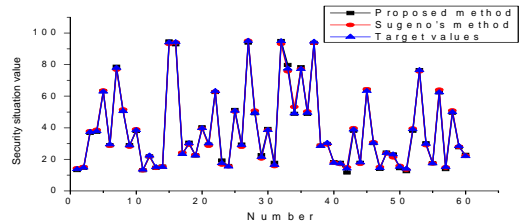

Figure 12. The experiment results of two method by number (t=32760)

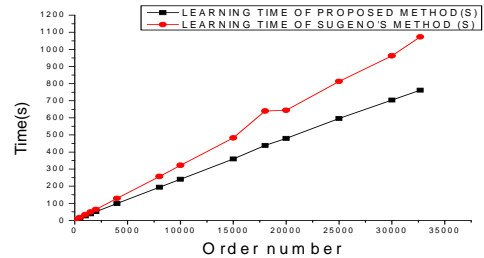

Figure 13. The comparison of learning time by order(s)

TABLE IV. COMPARISON RESULT OF LEARNING TIME AND ACCURACY FOR TWO METHODS (T=32760)

| No | Learning method | Time (s) | Accuracy (%) |
|---|---|---|---|
| 1 | Proposed method | 286 | 92.432 |
| 2 | Sugeno's method | 392.28 | 93.371 |
| 3 | update | 1.37times | -0.939 |

The neuron-fuzzy-based-obstacle avoidance program is simulated and implemented on the hardware system using Altera Quartus® II design software, System-on-programmable chip (SOPC) Builder, Nios® II Integrated Design Environment (IDE) software, and FPGA development and education board (DE2)[26].

There is the need that the learning algorithm is fast and has a small error on the mobile device using embedded system. Using DE2 board of Altera Corporation, the experiment is performed with *sugeno* method and proposed method (Fig. 14, Fig. 15).

Fig. 14 shows the comparison of learning time by order (precipitation forecast). Fig. 15 shows the comparison of learning time by order (security situation).

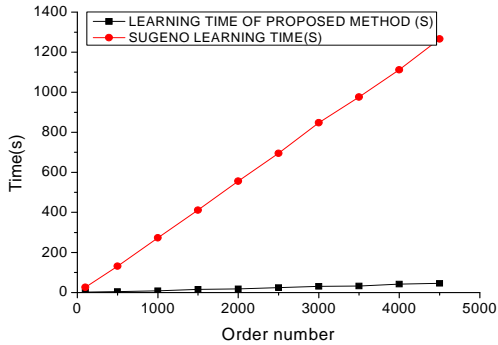

Figure 14. The comparison of precipitation learning time by order(s) (DE2)

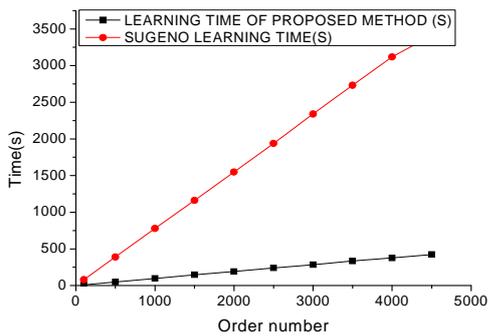

Figure 15. The comparison of security situation learning time by order(s) (DE2)

Table 5 shows average learning time result of two methods on DE2 board.

TABLE V. COMPARISON RESULT OF LEARNING TIME FOR TWO METHODS (DE2)

| learning time | Sugeno learning time /100 order | Learning time of proposed method /100 order | proportion (sugeno time/proposed time) |
|---|---|---|---|
| **precipitation** | 27.85 | 0.996 | 27.961 times |
| **security situation** | 79.812 | 9.535 | 8.370 times |

## V. ANALYSIS OF THE EXPERIMENT RESULT

The experiment of accuracy and time is studied while several programs are working at the same time. According to conditions of hardware, results of experiment are not same. But the time rate between methods is same. The conditions of hardware were CPU DELL 3 GHZ, Main board 915, Memory size 512 MB. The embedded system is the DE2 70 of Altera Corporation, chip Model is EP2C70F896C6, clock 100 M Hz.

The result shows, that parameter identification time for precipitation data is faster about 5.2 times and the parameter identification for security situation data is faster about 1.37 times compared with the existing *sugeno* method on PC. The result on DE2 board shows, the parameter identification time for precipitation data was faster about 27.961 times and the parameter identification for security situation data was faster about 8.37 times compared with the existing *sugeno* method. The *sugeno* method needs calculating of exponent function and DE2 has not exponent operation, So, two methods has big time difference. Therefore, on DE2 board, the parameter identification based on proposed method is very good. As for security situation, the accuracy became smaller because of the 3 variables. The time difference of the two methods will get bigger as fuzzy variables and rules are added. The identification accuracy of precipitation prediction has improved by 7.35%, but, the accuracy of security situation is decreased by 0.937%. The only reason was because fuzzy variables were 3 and the accuracy will get better if more fuzzy variables are added. The reason for time reduction is because the proposed method only uses the distance for the width of membership function and input information, the course of reasoning by using threshold is eliminated, even though *Sugeno* method uses the form of exponent function while fuzzy processing[27, 28, 29]. It can be easily known that the proposed method has higher efficiency in parameter identification compared with existing method.

## VI. CONCLUSION

This paper explained improved *T-S* fuzzy inference method based on moving rate and production term to improve the problem that the identification algorithm for fuzzy neural networks has big time complexity. The moving rate and production term is defined, and inference method using efficient inference parameter is proposed. The parameter identification algorithm based on production inference method is proposed. The advantages of the proposed inference method are proved by comparing with the existing *Sugeno* method. The efficiency of the proposed identification method is also proved that the learning time is faster at least 1.37 times and 8.37 times compared with existing *Sugeno* method through experiment of precipitation and security situation data on PC and DE2 FPGA board. The result of experiment shows that proposed method based on moving rate and production term has advantages of shorter time complexity and low learning error compared with the existing method. For future work, the structure identification method based on production inference should be considered.